%% file: main.tex
\documentclass[conference]{IEEEtran}
\usepackage{times}

\usepackage[numbers]{natbib}
\usepackage{multicol}
\usepackage[bookmarks=true]{hyperref}

\input{defines}

\begin{document}

\title{Contact-Aided Invariant Extended Kalman Filtering for Legged Robot State Estimation}

\author{
\authorblockN{Ross Hartley, Maani Ghaffari Jadidi, Jessy W. Grizzle, and Ryan M. Eustice}
\authorblockA{
College of Engineering, University of Michigan, Ann Arbor, MI, USA}
\tt\small \{{rosshart, maanigj, grizzle, eustice\}}@umich.edu
}
\maketitle

\begin{abstract}
This paper derives a contact-aided inertial navigation observer for a 3D bipedal robot using the theory of invariant observer design. Aided inertial navigation is fundamentally a nonlinear observer design problem; thus, current solutions are based on approximations of the system dynamics, such as an Extended Kalman Filter (EKF), which uses a system's Jacobian linearization along the current best estimate of its trajectory. On the basis of the theory of invariant observer design by Barrau and Bonnabel, and in particular, the Invariant EKF (InEKF), we show that the error dynamics of the point contact-inertial system follows a log-linear autonomous differential equation; hence, the observable state variables can be rendered convergent with a domain of attraction that is independent of the system's trajectory. Due to the log-linear form of the error dynamics, it is not necessary to perform a nonlinear observability analysis to show that when using an Inertial Measurement Unit (IMU) and contact sensors, the absolute position of the robot and a rotation about the gravity vector (yaw) are unobservable. We further augment the state of the developed InEKF with IMU biases, as the online estimation of these parameters has a crucial impact on system performance. We evaluate the convergence of the proposed system with the commonly used quaternion-based EKF observer using a Monte-Carlo simulation. In addition, our experimental evaluation using a Cassie-series bipedal robot shows that the contact-aided InEKF provides better performance in comparison with the quaternion-based EKF as a result of exploiting symmetries present in the system dynamics.
\end{abstract}

\IEEEpeerreviewmaketitle

\input{intro}
\input{preliminaries}
\input{riekf}

\input{bias}

\input{contact_switch}
\input{experimental_results}

\section{Conclusion} 
\label{sec:conclusion}
Using recent results on a group-invariant form of the extended Kalman filter (EKF), this article derived an observer for a contact-aided inertial navigation system for a 3D legged robot. Contact and IMU sensors are available on all modern bipedal robots; therefore, the developed system has the potential to become an essential part of such platforms since an observer with a large basin of attraction can improve the reliability of perception and control algorithms. We also included IMU biases in the state estimator and showed that, while some of the theoretical guarantees are lost, in real experiments, the proposed system has better performance than that of a commonly used quaternion-based EKF. Although the latter is a discrete EKF on Lie group, it does not exploit symmetries present in the system dynamics and observation models, namely, invariance of the estimation error under a group action. Future work includes integration of the observer developed in this work with a camera-based perception system for outdoor terrain mapping and navigation. 

\section*{Acknowledgments}
\small{
Funding for R. Hartley and M. Ghaffari Jadidi is given by the Toyota Research Institute (TRI), partly under award number N021515, however this article solely reflects the opinions and conclusions of its authors and not TRI or any other Toyota entity. Funding for J. Grizzle was in part provided by TRI and in part by NSF Award No.~1525006.}

\small
\bibliographystyle{plainnat}
\bibliography{references}

\end{document}

%% file: defines.tex
\usepackage{graphicx}
\usepackage{caption}
\graphicspath{
    {graphics/}
}

\usepackage{amsmath,amssymb,enumerate}

\usepackage{amsthm}
\usepackage{setspace}
\usepackage{booktabs}
\usepackage[usenames,dvipsnames,svgnames,table]{xcolor}
\usepackage{mathtools}
\usepackage{algorithm, algorithmicx, algpseudocode}
\usepackage{blindtext}
\usepackage{gensymb}
\usepackage{xparse}
\usepackage{lipsum}
\usepackage{graphicx}
\usepackage{soul} 
\usepackage{mathrsfs}
\usepackage[mathscr]{euscript}
\usepackage{times}
\usepackage[numbers]{natbib}
\usepackage{multicol}
\usepackage[caption=false,font=footnotesize]{subfig}
\usepackage{amsfonts}
\usepackage{fixltx2e}
\usepackage{cleveref}
\usepackage[utf8]{inputenc}
\usepackage[T1]{fontenc}
\usepackage{textcomp}

\newtheorem{theorem}{Theorem}

\newtheorem{remark}{Remark}

\newtheorem{definition}{Definition}

\def\realnumbers{\mathbb{R}}

\renewcommand{\thefigure}{\arabic{figure}}

\definecolor{note}{rgb}{0.1,0.1,1}
\definecolor{rephase}{rgb}{0.15,0.7,0.15}
\definecolor{bag}{rgb}{0.6,0.6,0.2}

\newcommand{\m}{\mathop{\mathrm{m}}}
\newcommand{\rad}{\mathop{\mathrm{rad}}}

\newcommand{\Hz}{\mathop{\mathrm{Hz}}}

\newcommand{\transpose}{\mathsf{T}}
\newcommand{\SO}{\mathrm{SO}}
\newcommand{\SE}{\mathrm{SE}}

\DeclareDocumentCommand{\vectorToSkew}{ O{} }{\left(#1\right)_\times}

\DeclareDocumentCommand{\vector}{ O{} }{\mathrm{vec}(#1)}
\DeclareDocumentCommand{\zeros}{ O{} }{\textbf{0}_{#1}}
\DeclareDocumentCommand{\X}{ O{} O{} }{\textbf{X}_{#1}^{#2}}
\DeclareDocumentCommand{\XE}{ O{} O{} }{\hat{\textbf{X}}_{#1}^{#2}}
\DeclareDocumentCommand{\L}{ O{} O{} }{\textbf{L}_{#1}^{#2}}
\DeclareDocumentCommand{\S}{ O{} O{} }{\textbf{S}_{#1}^{#2}}
\DeclareDocumentCommand{\P}{ O{} O{} }{\textbf{P}_{#1}^{#2}}
\DeclareDocumentCommand{\B}{ O{} O{} }{\textbf{B}_{#1}^{#2}}
\DeclareDocumentCommand{\Q}{ O{} O{} }{\textbf{Q}_{#1}^{#2}}
\DeclareDocumentCommand{\H}{ O{} O{} }{\textbf{H}_{#1}^{#2}}
\DeclareDocumentCommand{\J}{ O{} O{} }{\textbf{J}_{#1}^{#2}}
\DeclareDocumentCommand{\R}{ O{} O{} }{\textbf{R}_{#1}^{#2}}
\DeclareDocumentCommand{\RE}{ O{} O{} }{\hat{\textbf{R}}_{#1}^{#2}}
\DeclareDocumentCommand{\Q}{ O{} O{} }{\textbf{Q}_{#1}^{#2}}
\DeclareDocumentCommand{\T}{ O{} O{} }{\textbf{T}_{#1}^{#2}}
\DeclareDocumentCommand{\L}{ O{} O{} }{\textbf{L}_{#1}^{#2}}
\DeclareDocumentCommand{\K}{ O{} O{} }{\textbf{K}_{#1}^{#2}}
\DeclareDocumentCommand{\V}{ O{} O{} }{\textbf{V}_{#1}^{#2}}
\DeclareDocumentCommand{\N}{ O{} O{} }{\textbf{N}_{#1}^{#2}}
\DeclareDocumentCommand{\Y}{ O{} O{} }{\textbf{Y}_{#1}^{#2}}
\DeclareDocumentCommand{\F}{ O{} O{} }{\textbf{F}_{#1}^{#2}}
\DeclareDocumentCommand{\G}{ O{} O{} }{\textbf{G}_{#1}^{#2}}
\DeclareDocumentCommand{\A}{ O{} O{} }{\textbf{A}_{#1}^{#2}}
\DeclareDocumentCommand{\TH}{ O{} O{} }{\boldsymbol{\Theta}_{#1}^{#2}}
\DeclareDocumentCommand{\x}{ O{} O{} }{\textbf{x}_{#1}^{#2}}
\DeclareDocumentCommand{\e}{ O{} O{} }{\textbf{e}_{#1}^{#2}}
\DeclareDocumentCommand{\c}{ O{} O{} }{\textbf{c}_{#1}^{#2}}
\DeclareDocumentCommand{\C}{ O{} O{} }{\textbf{C}_{#1}^{#2}}
\DeclareDocumentCommand{\CM}{ O{} O{} }{\tilde{\textbf{C}}_{#1}^{#2}}
\DeclareDocumentCommand{\RM}{ O{} O{} }{\tilde{\textbf{R}}_{#1}^{#2}}
\DeclareDocumentCommand{\I}{ O{} O{} }{\textbf{I}_{#1}^{#2}}
\DeclareDocumentCommand{\O}{ O{} O{} }{\textbf{O}_{#1}^{#2}}
\DeclareDocumentCommand{\r}{ O{} O{} }{\textbf{r}_{#1}^{#2}}
\DeclareDocumentCommand{\t}{ O{} O{} }{\textbf{t}_{#1}^{#2}}
\DeclareDocumentCommand{\u}{ O{} O{} }{\textbf{u}_{#1}^{#2}}
\DeclareDocumentCommand{\d}{ O{} O{} }{\textbf{d}_{#1}^{#2}}
\DeclareDocumentCommand{\dE}{ O{} O{} }{\hat{\textbf{d}}_{#1}^{#2}}
\DeclareDocumentCommand{\b}{ O{} O{} }{\textbf{b}_{#1}^{#2}}
\DeclareDocumentCommand{\a}{ O{} O{} }{\textbf{a}_{#1}^{#2}}
\DeclareDocumentCommand{\g}{ O{} O{} }{\textbf{g}_{#1}^{#2}}
\DeclareDocumentCommand{\dM}{ O{} O{} }{\tilde{\textbf{d}}_{#1}^{#2}}
\DeclareDocumentCommand{\params}{ O{} O{} }{\boldsymbol{\theta}_{#1}^{#2}}
\DeclareDocumentCommand{\paramsE}{ O{} O{} }{\hat{\boldsymbol{\theta}}_{#1}^{#2}}
\DeclareDocumentCommand{\paramError}{ O{} O{} }{\boldsymbol{\zeta}_{#1}^{#2}}
\DeclareDocumentCommand{\gyroscopeBias}{ O{} }{\textbf{b}_{#1}^{g}}
\DeclareDocumentCommand{\gyroscopeBiasE}{ O{} }{\hat{\textbf{b}}_{#1}^{g}}
\DeclareDocumentCommand{\accelerometerBias}{ O{} }{\textbf{b}_{#1}^{a}}
\DeclareDocumentCommand{\accelerometerBiasE}{ O{} }{\hat{\textbf{b}}_{#1}^{a}}
\DeclareDocumentCommand{\position}{ O{} O{} }{{}_\text{#2}\textbf{p}_{\text{#1}}}
\DeclareDocumentCommand{\positionDot}{ O{} O{} }{{}_\text{#2}\dot{\textbf{p}}_{\text{#1}}}
\DeclareDocumentCommand{\linearVelocity}{ O{} O{} }{{}_\text{#2}\textbf{v}_{\text{#1}}}
\DeclareDocumentCommand{\linearVelocityM}{ O{} O{} }{{}_\text{#2}\tilde{\textbf{v}}_{\text{#1}}}
\DeclareDocumentCommand{\orientation}{ O{} }{\textbf{R}_{\text{#1}}}
\DeclareDocumentCommand{\orientationDot}{ O{} }{\dot{\textbf{R}}_{\text{#1}}}
\DeclareDocumentCommand{\angularVelocity}{ O{} O{} }{{}_\text{#2}\boldsymbol{\omega}_{\text{#1}}}
\DeclareDocumentCommand{\angularVelocityM}{ O{} O{} }{{}_\text{#2}\tilde{\boldsymbol{\omega}}_{\text{#1}}}
\DeclareDocumentCommand{\acceleration}{ O{} O{} }{{}_\text{#2}\textbf{a}_{\text{#1}}}
\DeclareDocumentCommand{\accelerationM}{ O{} O{} }{{}_\text{#2}\tilde{\textbf{a}}_{\text{#1}}}
\DeclareDocumentCommand{\p}{ O{} O{} }{\textbf{p}_{#1}^{#2}}
\DeclareDocumentCommand{\pE}{ O{} O{} }{\hat{\textbf{p}}_{#1}^{#2}}
\DeclareDocumentCommand{\pM}{ O{} O{} }{\tilde{\textbf{p}}_{#1}^{#2}}
\DeclareDocumentCommand{\v}{ O{} O{} }{\textbf{v}_{#1}^{#2}}
\DeclareDocumentCommand{\vE}{ O{} O{} }{\hat{\textbf{v}}_{#1}^{#2}}
\DeclareDocumentCommand{\w}{ O{} O{} }{\boldsymbol{\omega}_{#1}^{#2}}
\DeclareDocumentCommand{\wM}{ O{} O{} }{\tilde{\boldsymbol{\omega}}_{#1}^{#2}}
\DeclareDocumentCommand{\noise}{ O{} O{} }{\textbf{w}_{#1}^{#2}}
\DeclareDocumentCommand{\FK}{ O{} }{\;\textit{\textbf{h}}_{#1}}
\DeclareDocumentCommand{\FKswitch}{ O{} }{\;\textit{\textbf{g}}_{#1}}
\DeclareDocumentCommand{\angleTheta}{ O{} O{} }{\boldsymbol{\theta}_{#1}^{#2}}
\DeclareDocumentCommand{\anglePhi}{ O{} O{} }{\boldsymbol{\phi}_{#1}^{#2}}
\DeclareDocumentCommand{\encoders}{ O{} O{} }{\boldsymbol{\alpha}_{#1}^{#2}}
\DeclareDocumentCommand{\encodersM}{ O{} O{} }{\tilde{\boldsymbol{\alpha}}_{#1}^{#2}}
\DeclareDocumentCommand{\offsets}{ O{} O{} }{\boldsymbol{\epsilon}_{#1}^{#2}}
\DeclareDocumentCommand{\Cov}{ O{} O{} }{\boldsymbol{\Sigma}_{#1}^{#2}}
\DeclareDocumentCommand{\Axis}{ O{} O{} }{\textnormal{Axis}_{#1}^{#2}}
\DeclareDocumentCommand{\using}{ O{} O{} }{\stackrel{\mathmakebox[\widthof{=}]{\text{eq.} (#1)}}{#2} \enspace}
\DeclareDocumentCommand{\Adjoint}{ O{} }{\mathrm{Ad}_{#1}}
\DeclareDocumentCommand{\vectorToAlgebra}{ O{} }{\mathscr{L}_\mathfrak{g}\left(#1\right)}
\DeclareDocumentCommand{\groupError}{ O{} O{} }{\boldsymbol{\eta}_{#1}^{#2}}
\DeclareDocumentCommand{\tangentError}{ O{} O{} }{\boldsymbol{\xi}_{#1}^{#2}}

\newcommand{\squeezeup}{\vspace{-3mm}}

\makeatletter
\newcommand{\mathleft}{\@fleqntrue\@mathmargin0pt}
\newcommand{\mathcenter}{\@fleqnfalse}
\makeatother

%% file: intro.tex
\section{Introduction}
Legged robots often use nonlinear observers that fuse leg odometry and Inertial Measurement Unit (IMU) measurements to infer trajectory, controller inputs such as velocity, and calibration parameters~\citep{rotella2014state,benallegue2015estimation,eljaik2015multimodal,kuindersma2016optimization}. In view of a practical solution, designing a globally convergent observer is sacrificed for one with at best local properties, such as the Extended Kalman Filter (EKF)~\citep{grizzle1995extended,krener2003convergence,trawny2005indirect}. Furthermore, joint encoders and IMUs provide high frequency measurements which exacerbate the challenge of meeting rigorous real-time performance requirements in legged robots that arise from their direct and time-varying contact with the environment~\cite{bloesch2013state,fankhauser2014robot,bloesch2017state,fallon2014drift,nobiliheterogeneous}. 

The theory of invariant observer design is based on the estimation error being invariant under the action of a matrix Lie group~\citep{aghannan2002invariant,bonnabel2009non}, which has recently led to the development of the Invariant EKF (InEKF)~\citep{bonnabel2007left,barrau2015non,barrau2017invariant,barrau2018invariant} with successful applications and promising results in simultaneous localization and mapping~\citep{barrau2015non,zhang2017convergence} and aided inertial navigation systems~\citep{barczyk2011invariant,barczyk2013invariant,barrau2015non,wu2017invariant}. The invariance of the estimation error with respect to a Lie group action is referred to as the symmetries of the system~\citep{barczyk2013invariant}. The main result of the InEKF is that symmetries lead to the estimation error satisfying a ``log-linear'' autonomous differential equation on the Lie algebra of the corresponding Lie group of system dynamics. Therefore, one can design a nonlinear observer or state estimator with strong convergence properties, which is rare.


\begin{figure} 
  \centering
  	\vspace{0.5em}
     \includegraphics[width=0.99\columnwidth]{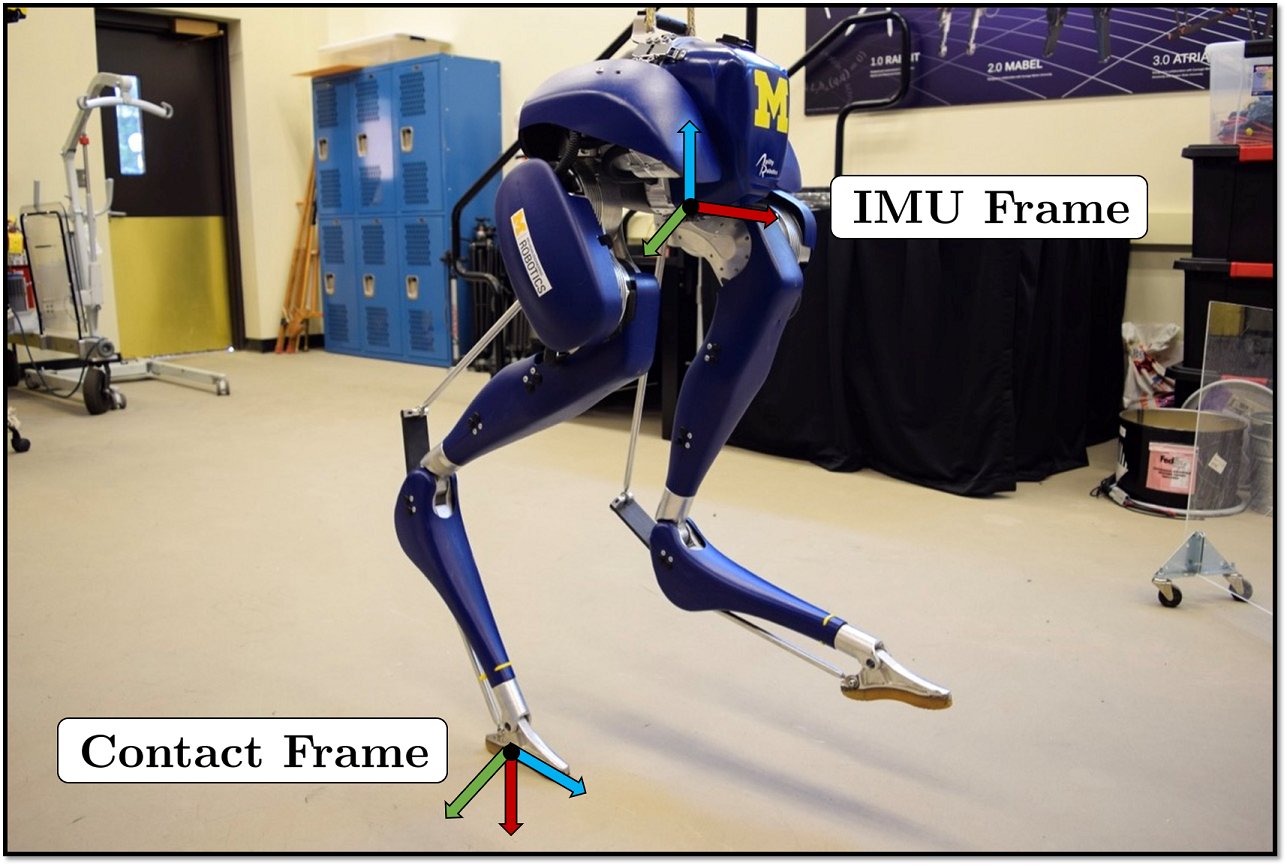}
    \caption{A Cassie-series biped robot is used for both simulation and experimental results. The robot was developed by Agility Robotics and has 20 degrees of freedom, 10 actuators, joint encoders, and an inertial measurement unit (IMU). The contact and IMU frames used in this work are depicted above.}
\label{fig:cassie}
\squeezeup\squeezeup
\end{figure}



In this article, we derive an InEKF for arbitrary matrix Lie groups acting on a system containing an IMU and contact sensor dynamics, and forward kinematics (FK) measurements. We show that the defined system satisfies the ``group affine'' property (log-linear error dynamics) and, therefore, can be incorporated as the process model of an InEKF. We further discuss inclusion of IMU bias into the observer which is necessary for real-world applications. This work has the following contributions:

\begin{enumerate}
\item Derivation of a right-invariant EKF for IMU and contact process model with a FK measurement model; the observability analysis is also presented;
\item State augmentation of above observer with IMU biases;
\item Evaluations of the derived observers in simulation and hardware experiments using a 3D bipedal robot;
\item An open-source implementation of the developed invariant observers can be found at \url{https://github.com/UMich-BipedLab/Contact-Aided-Invariant-EKF}.
\end{enumerate}

The remainder of this article is organized as follows. Background and preliminaries are given in Section~\ref{sec:prelim}. Section~\ref{sec:riekf} provides the derivation of a right-invariant EKF for contact-inertial navigation with a right-invariant FK measurement model. We also present simulation results of the convergence. Section~\ref{sec:bias} discusses the state augmentation of the previously derived InEKF with IMU bias. The consequences of the switching contact points on the state estimator are described in Section~\ref{sec:switchcontact}. Experimental evaluations on a 3D biped robot are presented in Section~\ref{sec:experimental_results}. Finally, Section~\ref{sec:conclusion} concludes the article and suggests future directions.

%% file: preliminaries.tex
\vspace{-.09cm}
\section{Review of Theoretical Background and Preliminaries}
\label{sec:prelim}
We assume a matrix Lie group~\citep{hall2015lie,chirikjian2011stochastic} denoted $\mathcal{G}$ and its associated Lie Algebra denoted $\mathfrak{g}$. If elements of $\mathcal{G}$ are $n \times n$ matrices, then so are elements of $\mathfrak{g}$. When doing calculations, it is very convenient to let
$$\mathcal{L}_\mathfrak{g}:\realnumbers^{\mathrm{dim} \mathfrak{g}} \to \mathfrak{g}$$
be the linear map that takes elements of the tangent space of $\mathcal{G}$ at the identity to the corresponding matrix representation so that the \textit{exponential map of the Lie group}, $\textnormal{exp}:\realnumbers^{\mathrm{dim} \mathfrak{g}} \to \mathcal{G}$,
is computed by
 $$\textnormal{exp}(\tangentError) = \textnormal{exp}_m(\vectorToAlgebra[\tangentError]),$$ where $\textnormal{exp}_m(\cdot)$ is the usual exponential of $n \times n$ matrices.

A process dynamics evolving on the Lie group with state at time $t$, $\X[t] \in \mathcal{G}$, is denoted by
$$\dfrac{\mathrm{d}}{\mathrm{d}t} \X[t] = f_{u_t}(\X[t]), $$
and $\XE[t]$ is used to denote an estimate of the state. The state estimation error is defined using right or left multiplication of $\X[t][-1]$  as follows.
\begin{definition}[Left and Right Invariant Error] 
The right- and left-invariant errors between two trajectories $\X[t]$ and $\XE[t]$ are:
\begin{equation}
\begin{split}
\groupError[t][r] &= \XE[t] \X[t][-1] = (\XE[t] \textbf{L}) (\X[t] \textbf{L})^{-1} \quad \text{(Right-Invariant)}\\
\groupError[t][l] &= \X[t][-1] \XE[t] = (\textbf{L} \XE[t])^{-1} (\textbf{L} \X[t]), \quad \text{(Left-Invariant)}
\end{split}
\end{equation}
where $\textbf{L}$ is an arbitrary element of the group.
\end{definition}
The following two theorems are the fundamental results for deriving an InEKF and show that by correct parametrization of the error variable,  a wide range of nonlinear problems can lead to linear error equations.
\begin{theorem}[Autonomous Error Dynamics \cite{barrau2017invariant}] \label{theorem:autonomous_error_dynamics}
A system is group affine if the dynamics, $f_{u_t}(\cdot)$, satisfies:
\begin{equation} \label{eq:group_affine}
f_{u_t}(\textbf{X}_1 \textbf{X}_2) = f_{u_t}(\textbf{X}_1) \textbf{X}_2 + \textbf{X}_1 f_{u_t}(\textbf{X}_2) - \textbf{X}_1 f_{u_t}(\I[d]) \textbf{X}_2
\end{equation}
for all $t>0$ and $\textbf{X}_1, \textbf{X}_2 \in \mathcal{G}$. Furthermore, if this condition is satisfied, the right- and left-invariant error dynamics are trajectory independent and satisfy:
\begin{alignat*}{2}
\dfrac{\mathrm{d}}{\mathrm{d}t} \groupError[t][r] &= g_{u_t}(\groupError[t][r]) \quad \text{where} \quad
g_{u_t}(\groupError[][r]) &&= f_{u_t}(\groupError[][r]) - \groupError[][r] f_{u_t}(\I[d])  \\
\dfrac{\mathrm{d}}{\mathrm{d}t} \groupError[t][l] &= g_{u_t}(\groupError[t][l]) \quad \text{where} \quad
g_{u_t}(\groupError[][l]) &&=  f_{u_t}(\groupError[][l]) - f_{u_t}(\I[d]) \groupError[][l] \\
\end{alignat*}
\end{theorem}
In the above, $\I[d] \in \mathcal{G}$ denotes the group identity element; to avoid confusion, we use $\I$ for a $3 \times 3$ identity matrix, and $\I[n]$ for the $n \times n$ case. In the following, for simplicity, we will use only the right-invariant error dynamics. 

Define $\A[t]$ to be a $\mathrm{dim} \mathfrak{g} \times \mathrm{dim} \mathfrak{g}$ matrix satisfying 
$$g_{u_t}(\textnormal{exp}(\tangentError[])) \triangleq \vectorToAlgebra[\A[t] \tangentError[]] + \mathcal{O}(||\tangentError[]||^2).$$ 
For all $t \ge 0$, let $\tangentError[t]$ be the solution of  the linear differential equation
\begin{equation}
\label{eq:LTVODE}
\dfrac{\mathrm{d}}{\mathrm{d}t} \tangentError[t][] = \A[t] \tangentError[t][].
\end{equation}

\begin{theorem}[Log-Linear Property of the Error \cite{barrau2017invariant}] \label{theorem:log_linear_error}
Consider the right-invariant error, $\groupError[t][]$, between two trajectories (possibly far apart). 
For arbitrary initial error $\tangentError[0][] \in \realnumbers^{\mathrm{dim} \mathfrak{g}}$, if 
\mbox{$\groupError[0][] =\exp(\tangentError[0][])$}, then for all $t\ge 0$, 
\begin{equation*}
\groupError[t][] = \textnormal{exp}(\tangentError[t][]);
\end{equation*}
that is, the nonlinear estimation error $\groupError[t][]$ can be exactly recovered from the time-varying linear differential equation \eqref{eq:LTVODE}.
\end{theorem}
This theorem states that \eqref{eq:LTVODE} is not the typical Jacobian linearization along a trajectory because the (left- or) right-invariant error on the Lie group can be exactly recovered from its solution. This result is of major importance for the propagation (prediction) step of the InEKF~\cite{barrau2017invariant}.

The adjoint representation plays a key role in the theory of Lie groups and through this linear map we can capture the non-commutative structure of a Lie group.
   \begin{definition}[The Adjoint Map, see page 63~\citet{hall2015lie}]
Let $\mathcal{G}$ be a matrix Lie group with Lie algebra $\mathfrak{g}$. For any $\textbf{X} \in \mathcal{G}$ the adjoint map, $\mathrm{Ad}_\textbf{X}:\mathfrak{g} \to \mathfrak{g}$, is a linear map defined as \mbox{$\mathrm{Ad}_\textbf{X}(\vectorToAlgebra[\tangentError]) = \textbf{X} \vectorToAlgebra[\tangentError] \textbf{X}^{-1}$}. Furthermore, we denote the matrix representation of the adjoint map by $\Adjoint[\X]$.
\end{definition}

For more details on the material discussed above, we refer reader to~\citet{barrau2015non,barrau2017invariant,barrau2018invariant}.

%% file: riekf.tex
\section{$\SE_{N+2}(3)$ Continuous Right-Invariant EKF}
\label{sec:riekf}
In this section, we derive a Right-Invariant Extended Kalman Filter (RI-EKF) using IMU and contact motion models with corrections made through forward kinematic measurements. This RI-EKF can be used to estimate the state of a robot that has an arbitrary (finite) number of points in contact with the static environment. While the filter is particularly useful for legged robots, the same theory can be applied for manipulators as long as the contact assumptions (presented in Section \ref{sec:riekf_dynamics}) are verified. 

In order to be consistent with the standard InEKF theory, IMU biases are neglected for now. Section \ref{sec:bias} provides a method for reintroducing the bias terms.

\subsection{State Representation}
As with typical aided inertial navigation, we wish to estimate the orientation, velocity, and position of the IMU (body) in the world frame \cite{lupton2012visual,forster2016manifold,yang2017monocular}. These states are represented by $\orientation[WB](t), \linearVelocity[WB][W](t)$, and $\position[WB][W](t)$ respectively. In addition, we append the position of all contact points (in the world frame), ${}_\text{W}\p[\text{WC}_i](t)$, to the list of state variables. This is similar to the approach taken in \cite{bloesch2013state,bloesch2017state}.

The above collection of state variables forms a matrix Lie group, $\mathcal{G}$. Specifically, for $N$ contact points, $\X[t] \in \SE_{N+2}(3)$ can be represented by the following matrix:
\begin{equation*}
\resizebox{\hsize}{!}{$
\nonumber \textbf{X}_t \triangleq
\begin{bmatrix}
\orientation[WB](t) & \linearVelocity[WB][W](t) & \position[WB][W](t) & {}_\text{W}\p[\text{WC}_1](t) & \cdots & {}_\text{W}\p[\text{WC}_N](t)  \\
\zeros[1,3] & 1 & 0 & 0 & \cdots & 0 \\
\zeros[1,3] & 0 & 1 & 0 & \cdots & 0 \\
\zeros[1,3] & 0 & 0 & 1 & \cdots & 0 \\
\vdots & \vdots & \vdots & \vdots & \ddots & \vdots \\ 
\zeros[1,3] & 0 & 0 & 0 & \cdots & 1 \\
\end{bmatrix}$}
\end{equation*}
Because the process and measurements models for each contact point, ${}_\text{W}\p[\text{WC}_i](t)$, are identical, without loss of generality, we will derive all further equations assuming only a single contact point. Furthermore, for the sake of readability, we introduce the following shorthand notation:
\begin{equation*}
\begin{split}
\X[t] \triangleq
\begin{bmatrix}
\R[t] & \v[t] & \p[t] & \d[t] \\
\zeros[1,3] & 1 & 0 & 0 \\
\zeros[1,3] & 0 & 1 & 0 \\
\zeros[1,3] & 0 & 0 & 1 \\
\end{bmatrix}
,\quad
\u[t] = 
\begin{bmatrix}
\angularVelocityM[WB][B](t) \\
\accelerationM[WB][B](t) \\
\end{bmatrix} 
\triangleq 
\begin{bmatrix}
\angularVelocityM[t] \\
\accelerationM[t] \\
\end{bmatrix}, 
\end{split}
\end{equation*}
where the input $\u[t]$ is formed from the angular velocity and linear acceleration measurements coming from the IMU. It is important to note that these measurements are taken in the body (or IMU) frame.
The Lie algebra of $\mathcal{G}$, denoted by $\mathfrak{g}$, is an $N+5$ dimensional square matrix. Following \cite{barrau2017invariant}, define a map, $\mathcal{L}_\mathfrak{g}: \realnumbers^{3N+9} \to \mathfrak{g}$, that maps a vector to the corresponding element of the Lie algebra. In the case of a single contact, for example, this function is defined by: 
\begin{equation}
\nonumber \vectorToAlgebra[\tangentError] = 
\begin{bmatrix}
\vectorToSkew[\tangentError[][R]] & \tangentError[][v] & \tangentError[][p] & \tangentError[][d] \\
\zeros[1,3] & 0 & 0 & 0 \\
\zeros[1,3] & 0 & 0 & 0 \\
\zeros[1,3] & 0 & 0 & 0 \\
\end{bmatrix},
\end{equation}
where $\vectorToSkew[\cdot]$ denotes a $3\times 3$ skew-symmetric matrix. The adjoint operator is given by:
\begin{equation*}
\Adjoint[\X[t]] = 
\begin{bmatrix}
\R & \zeros & \zeros & \zeros \\
\vectorToSkew[\v[t]] \R[t] & \R[t] & \zeros & \zeros \\
\vectorToSkew[\p[t]] \R[t] & \zeros & \R[t] & \zeros \\
\vectorToSkew[\d[t]] \R[t] & \zeros & \zeros & \R[t] \\
\end{bmatrix}.
\end{equation*}

\subsection{Continuous System Dynamics} \label{sec:riekf_dynamics}
The IMU measurements are modeled as being corrupted by additive white Gaussian noise, per
\begin{alignat*}{2}
\angularVelocityM[t] &= \angularVelocity[t] + \noise[t][g], \qquad &&\noise[t][g] \sim \mathcal{N}(\zeros[3,1], \Cov[][g]) \\
\accelerationM[t] &= \acceleration[t] + \noise[t][a], \qquad &&\noise[t][a] \sim \mathcal{N}(\zeros[3,1], \Cov[][a]); 
\end{alignat*}
these are explicit measurements coming directly from a physical sensor. In contrast, the velocity of the contact point is implicitly \textit{inferred} through a contact sensor; specifically, when a binary sensor indicates contact, the position of the contact point is \textit{assumed to remain fixed in the world frame}, i.e. the measured velocity is zero. In order to accommodate potential slippage, the measured velocity is assumed to be the actual velocity plus white Gaussian noise, namely
\begin{equation*}
\linearVelocityM[WC][C] = \zeros[3,1] = \linearVelocity[WC][C] + \noise[t][v], \quad \noise[t][v] \sim \mathcal{N}(\zeros[3,1], \Cov[][v]). \\
\end{equation*}
Using the IMU and contact measurements, the individual terms of the system dynamics can be written as:
\begin{equation}
\begin{split}
\dot{\textbf{R}}_t &= \R[t]\vectorToSkew[\angularVelocityM[t] - \noise[t][g]] \\ 
\dot{\textbf{v}}_t &= \R[t](\accelerationM[t] - \noise[t][a]) + \g \\
\dot{\textbf{p}}_t &= \v[t] \\ 
\dot{\textbf{d}}_t &= \R[t] \FK[R](\encodersM[t])(-\noise[t][v]),
\end{split}
\end{equation}
where $\g$ is the gravity vector and $\FK[R](\encodersM[t])$, arising from forward kinematics, is the measured orientation of the contact frame with respect to the IMU frame. Therefore, $\R[t] \FK[R](\encodersM[t])$ is a rotation matrix that transforms a vector from the contact frame to the world frame.

In matrix form, the dynamics can be expressed as
\begin{equation*}
\begin{split}
&\dfrac{\mathrm{d}}{\mathrm{d}t} \X[t] = 
\begin{bmatrix}
  \R[t]\vectorToSkew[\angularVelocityM[t]] 
& \R[t]\accelerationM[t] + \g  
& \v[t]
& \zeros[3,1] \\
\zeros[1,3] & 0 & 0 & 0  \\
\zeros[1,3] & 0 & 0 & 0  \\
\zeros[1,3] & 0 & 0 & 0  \\
\end{bmatrix} \\
&- 
\begin{bmatrix}
\R[t] & \v[t] & \p[t] & \d[t] \\
\zeros[1,3] & 1 & 0 & 0 \\
\zeros[1,3] & 0 & 1 & 0 \\
\zeros[1,3] & 0 & 0 & 1 \\
\end{bmatrix}
\begin{bmatrix}
\vectorToSkew[\noise[t][g]] 
& \noise[t][a]
& \zeros[3,1]
& \FK[R](\encodersM[t]) \noise[t][v]  \\
\zeros[1,3] & 0 & 0 & 0  \\
\zeros[1,3] & 0 & 0 & 0  \\
\zeros[1,3] & 0 & 0 & 0  \\
\end{bmatrix} \\
&\triangleq f_{u_t}(\X[t]) - \X[t] \vectorToAlgebra[\noise[t]],
\end{split}
\end{equation*}
with \mbox{$\noise[t] \triangleq \vector[\noise[t][g],\; \noise[t][a],\; \zeros[3,1], \FK[R](\encodersM[t]) \noise[t][v]]$}. The deterministic system dynamics, $f_{u_t}(\cdot)$, can be shown to satisfy the group affine property, \eqref{eq:group_affine}. Therefore, following Theorem~\ref{theorem:autonomous_error_dynamics}, the left- and right-invariant error dynamics will evolve independently of the system's state.

Using Theorem~\ref{theorem:autonomous_error_dynamics}, the right-invariant error dynamics is 
\begin{equation}
\begin{split}
\nonumber \dfrac{\mathrm{d}}{\mathrm{d}t} \groupError[t][r] &= f_{u_t}(\groupError[t][r]) - \groupError[t][r] f_{u_t}(\I[d]) + (\XE[t]  \vectorToAlgebra[\noise[t]] \XE[t][-1]) \groupError[t][r] \\
&\triangleq g_{u_t}(\groupError[t][r]) + \vectorToAlgebra[\hat{\noise}_t] \groupError[t][r] \\
\end{split}
\end{equation}
where the second term arises from the additive noise. The derivation follows the results in \cite{barrau2017invariant} and is not repeated here.

Theorem \ref{theorem:log_linear_error} furthermore, specifies that the invariant error satisfies a log-linear property. Namely, if $\A[t]$ is defined by $g_{u_t}(\text{exp}(\tangentError)) \triangleq \vectorToAlgebra[\A[t] \tangentError] + \mathcal{O}(||\tangentError||^2)$, then the log of the invariant error, $\tangentError \in \realnumbers^{\mathrm{dim} \mathfrak{g}}$, satisfies the linear system
\begin{equation} \label{eq:tangent_error_dynamics}
\begin{split}
\dfrac{\mathrm{d}}{\mathrm{d}t} \tangentError[t] &= \A[t] \tangentError[t] + \hat{\noise}_t = \A[t] \tangentError[t] + \Adjoint[\XE[t]] \noise[t] \\
\groupError[t][r] &= \text{exp}(\tangentError[t]).
\end{split}
\end{equation}
To compute the matrix $\A[t]$, we linearize the invariant error dynamics, $g_{u_t}(\cdot)$, using the first order approximation \mbox{$\groupError[t][r] = \text{exp}(\tangentError[t]) \approx \I[d] + \vectorToAlgebra[\tangentError[t]]$} to yield
\begin{equation} \label{eq:linear_error_dynamics}
\small
\begin{aligned} 
&g_{u_t}(\I[d] + \mathscr{L}_\mathfrak{g}(\boldsymbol{\xi})) = \\
&\begin{bmatrix}
\left( \I+ \vectorToSkew[\tangentError[t][R]] \right) \vectorToSkew[\angularVelocityM[t]] 
&\left( \I+ \vectorToSkew[\tangentError[t][R]] \right) \accelerationM[t] + \g & \tangentError[t][v] 
& \zeros[3,1]  \\
\zeros[1,3] & 0 & 0 & 0  \\
\zeros[1,3] & 0 & 0 & 0  \\
\zeros[1,3] & 0 & 0 & 0  \\
\end{bmatrix} \\
&-
\begin{bmatrix}
\I + \vectorToSkew[\tangentError[t][R]] & \tangentError[t][v] & \tangentError[t][p] & \tangentError[t][d] \\
\zeros[3,1] & 1 & 0 & 0 \\
\zeros[3,1] & 0 & 1 & 0  \\
\zeros[3,1] & 0 & 0 & 1 \\
\end{bmatrix}
\begin{bmatrix}
\vectorToSkew[\angularVelocityM[t]] & \accelerationM + \g & \zeros[3,1] & \zeros[3,1]  \\
\zeros[1,3] & 0 & 0 & 0  \\
\zeros[1,3] & 0 & 0 & 0  \\
\zeros[1,3] & 0 & 0 & 0  \\
\end{bmatrix} \\
&= 
\begin{bmatrix}
\zeros[3,3] & \vectorToSkew[\g] \tangentError[t][R]  & \tangentError[t][v] & \zeros[3,1] \\
\zeros[1,3] & 0 & 0 & 0  \\
\zeros[1,3] & 0 & 0 & 0  \\
\zeros[1,3] & 0 & 0 & 0  \\
\end{bmatrix}
= \vectorToAlgebra[
\begin{bmatrix} 
\zeros[3,1] \\
\vectorToSkew[\g] \tangentError[t][R] \\
\tangentError[t][v] \\
\zeros[3,1]
\end{bmatrix} ].
\end{aligned}
\end{equation}

With the above, we can express the prediction step of the RI-EKF. The state estimate, $\XE[t]$, is propagated though the deterministic system dynamics, while the covariance matrix, $\P[t]$, is computed using the Riccati equation, namely,
\begin{align}
\dfrac{\mathrm{d}}{\mathrm{d}t} \XE[t] = f_{u_t}(\XE[t])~\text{and}~
\dfrac{\mathrm{d}}{\mathrm{d}t} \P[t] = \A[t] \P[t] + \P[t] \A[t][\transpose] + \hat{\Q}_t,
\end{align}
where the matrices $\A[t]$ and $\hat{\Q}_t$ are obtained from \eqref{eq:linear_error_dynamics} and \eqref{eq:tangent_error_dynamics},
\begin{align}
\label{eq:firstAt}
\A[t] = 
\begin{bmatrix}
\zeros & \zeros & \zeros & \zeros  \\
\vectorToSkew[\g] & \zeros & \zeros & \zeros  \\
\zeros & \I & \zeros & \zeros \\
\zeros & \zeros & \zeros & \zeros   
\end{bmatrix} \text{and}~
\hat{\Q}_t = \Adjoint[\XE[t]] \text{Cov}\left( \noise[t] \right) \Adjoint[\XE[t]]^\transpose.
\end{align}
\begin{remark}
In~\eqref{eq:firstAt}, $\A[t]$ is time-invariant and the time subscript could be dropped. However, in general it can be time-varying, therefore, we use $\A[t]$ throughout the paper.
\end{remark}

\subsection{Right-invariant Forward Kinematic Measurement Model}
Let $\encoders[t] \in \realnumbers^M$ denote the vector of joint positions (prismatic or revolute) between the body and the contact point. We assume that the encoder measurements are corrupted by additive white Gaussian noise.
\begin{equation}
\encodersM[t] = \encoders[t] + \noise[t][\alpha], \quad \noise[t][\alpha] \sim \mathcal{N}(\zeros[M,1], \Cov[][\alpha])
\end{equation}
Using forward kinematics, we determine the relative position of the contact point with respect to the body,
\begin{equation}
\position[BC][B](t) \triangleq \FK[p](\encodersM[t] - \noise[t][\alpha]) \approx \FK[p](\encodersM[t]) - \J[v](\encodersM[t]) \noise[t][\alpha],
\end{equation}
where $\J[v]$ denotes the components of the geometric (``manipulator'') Jacobian corresponding to linear velocity \cite{murray2017mathematical}. Using the state variables, the forward-kinematics position measurement becomes
\begin{align} \label{eq:fk_measurement}
\FK[p](\encodersM[t]) = \R[t][\transpose](\d[t] - \p[t]) + \J[v](\encodersM[t]) \noise[t][\alpha].
\end{align}
Re-written in matrix form, this measurement has the right-invariant observation structure defined in \cite{barrau2017invariant}, $\Y[t] = \X[t][-1] \b + \V[t]$,
\begin{equation*} 
\resizebox{\hsize}{!}{$
\begin{bmatrix}
\FK[p](\encodersM[t]) \\
0 \\
1 \\
-1
\end{bmatrix} 
= 
\begin{bmatrix}
\R[t][\transpose] & -\R[t][\transpose]\v[t] & -\R[t][\transpose]\p[t] & -\R[t][\transpose]\d[t] \\
\zeros[1,3] & 1 & 0 & 0 \\
\zeros[1,3] & 0 & 1 & 0 \\
\zeros[1,3] & 0 & 0 & 1 \\
\end{bmatrix}
\begin{bmatrix}
\zeros[3,1] \\
0 \\
1 \\
-1
\end{bmatrix} 
+
\begin{bmatrix}
\J[v](\encodersM[t]) \noise[t][\alpha] \\
0 \\
0 \\
0 \\
\end{bmatrix}$}.
\end{equation*}
Therefore, the innovation depends solely on the invariant error and the update equations take the form \cite[Section 3.1.2]{barrau2017invariant}
\begin{align} \label{eq:riekf_update}
\begin{split}
\XE[t][+] &= \text{exp}\left( \L[t] \left( \XE[t] \Y[t] - \b \right) \right) \XE[t] \\
\groupError[t][r+] &= \text{exp}\left( \L[t] \left( \groupError[t][r] \b - \b + \XE[t] \V[t] \right)  \right) \groupError[t][r],  \\
\end{split}
\end{align}
where $\text{exp}(\cdot)$ is the exponential map corresponding to the state matrix Lie group, $\mathcal{G}$, $\L[t]$ is a gain matrix to be defined later, \mbox{\small$\b[][\transpose] = \begin{bmatrix} \zeros[1,3] & 0 & 1 & -1 \end{bmatrix}$}, and \mbox{\small$\Y[t][\transpose] = \begin{bmatrix} \FK[p]^\transpose(\encodersM[t]) & 0 & 1 & -1 \end{bmatrix}$}. Because the last three rows of \mbox{\small$\XE[t] \Y[t] - \b$} are identically zero, we can express the update equations using a reduced dimensional gain, $\K[t]$, and an auxiliary matrix \mbox{\small$\boldsymbol{\Pi} \triangleq \begin{bmatrix} \I & \zeros[3,3] \end{bmatrix}$}, so that \mbox{\small$\L[t] \left( \XE[t] \Y[t] - \b \right) = \K[t] \boldsymbol{\Pi} \left( \XE[t] \Y[t] \right)$} as detailed in \cite{barrau2015non}.

\begin{figure*}[t!]
  \centering
    \includegraphics[width=\textwidth]{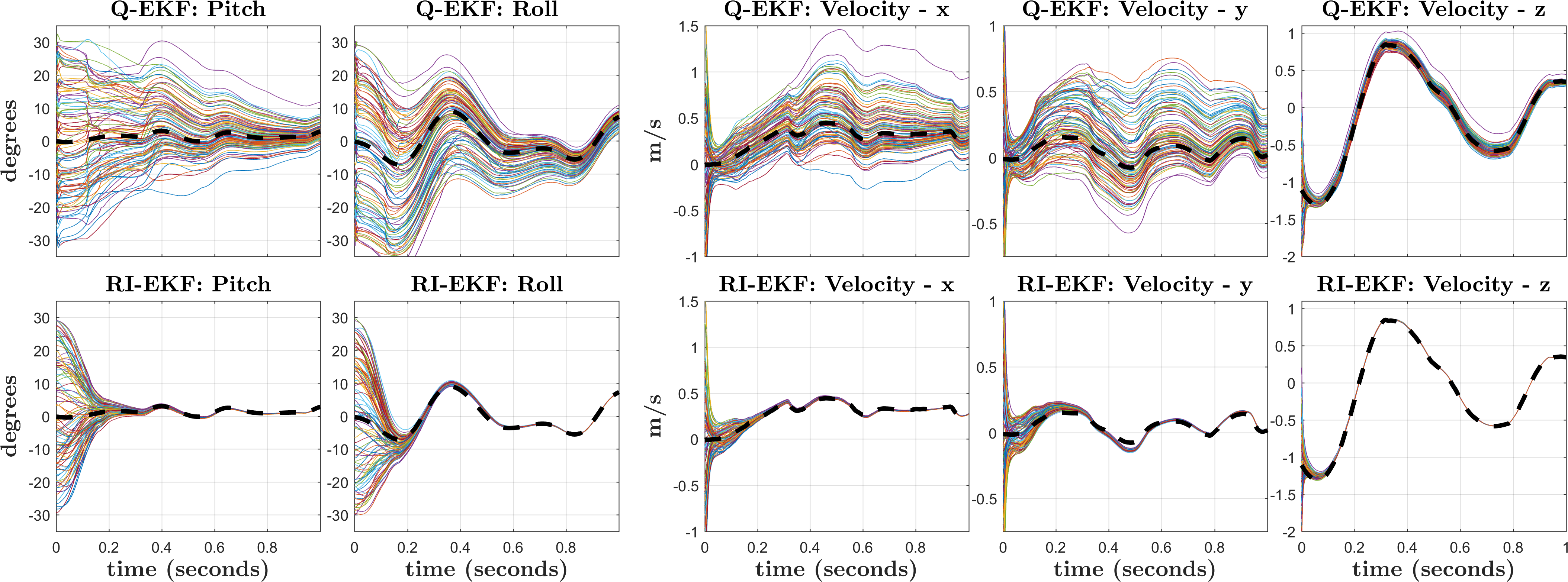}
      \caption{A quaternion-based EKF (Q-EKF) and the proposed right-invariant EKF (RI-EKF) were run 100 times using the same measurements, noise statistics, and initial covariance, but with random initial orientations and velocities. The noisy measurements came from a dynamic simulation of a Cassie-series biped robot where the robot walks forwards after a small drop, accelerating from $0.0$ to $0.3~\m/\sec$. The above plots show the state estimate for the first second of data, where the dashed black line represents the true state. The RI-EKF (bottom row) converges considerably faster than the Q-EKF (top row) for all observable states. The estimated yaw angle (not shown) does not converge for either filter because it is unobservable. Therefore, to compare convergence, the velocities shown are represented in the estimated IMU (body) frame.}
\label{fig:sim_comparison}
\squeezeup\squeezeup
\end{figure*}

Using the first order approximation of the exponential map, \mbox{\small$\groupError[t][r] = \text{exp}(\tangentError[t]) \approx \I[d] + \vectorToAlgebra[\tangentError[t]]$}, and dropping higher-order terms, we can linearize the update equation \eqref{eq:riekf_update},
\begin{equation*}
\small
\begin{aligned}
& \groupError[t][r+] \approx \I[d] + \vectorToAlgebra[\tangentError[t][+]] \approx \I[d] + \vectorToAlgebra[\tangentError[t]] \\
&+ \vectorToAlgebra[{ \K[t] \boldsymbol{\Pi} \left( \left(\I[d] + \vectorToAlgebra[\tangentError[t]]\right) 
\begin{bmatrix}
\zeros[3,1] \\ 0 \\ 1 \\ -1 
\end{bmatrix} 
+ \XE[t]
\begin{bmatrix}
\J[v](\encodersM[t]) \noise[t][\alpha] \\ 0 \\ 0 \\ 0
\end{bmatrix}
\right) }]. 
\end{aligned}
\end{equation*}
Therefore,
\begin{align*}
\small
\begin{split}
& \vectorToAlgebra[\tangentError[t][+]] = \vectorToAlgebra[\tangentError[t]] + \\ & \scriptsize{\vectorToAlgebra[{ 
\K[t] \boldsymbol{\Pi} \left( 
\begin{bmatrix}
\I + \vectorToSkew[\tangentError[t][R]] & \tangentError[t][v] & \tangentError[t][p] & \tangentError[t][d] \\
\zeros[3,1] & 1 & 0 & 0 \\
\zeros[3,1] & 0 & 1 & 0  \\
\zeros[3,1] & 0 & 0 & 1 \\
\end{bmatrix}
\begin{bmatrix}
\zeros[3,1] \\ 0 \\ 1 \\ -1 
\end{bmatrix}
+ \XE[t]
\begin{bmatrix}
\J[v](\encodersM[t]) \noise[t][\alpha] \\ 0 \\ 0 \\ 0
\end{bmatrix}
\right) }]} \\
& =  \vectorToAlgebra[\tangentError[t]] + \vectorToAlgebra[{ \K[t] \boldsymbol{\Pi} \left(
\begin{bmatrix}
\tangentError[t][p] - \tangentError[t][d] \\ 0 \\ 1 \\ -1 
\end{bmatrix}
+ \XE[t]
\begin{bmatrix}
\J[v](\encodersM[t]) \noise[t][\alpha] \\ 0 \\ 0 \\ 0
\end{bmatrix}
\right) }].
\end{split}
\end{align*}
Taking $\mathcal{L}_\mathfrak{g}^{-1}$ of both sides yields the linear update equation:
\begin{align} \label{eq:linear_update}
\nonumber \tangentError[t][+] &= \tangentError[t] - \K[t] \left(
\begin{bmatrix}
\zeros[3,3] & \zeros[3,3] & -\I & \I
\end{bmatrix}
\tangentError[t] - \hat{\R}_t (\J[v](\encodersM[t]) \noise[t][\alpha])
\right) \\
&\triangleq \tangentError[t] - \K[t] \left( \H[t] \tangentError[t] - \hat{\R}_t \left(\J[v](\encodersM[t]) \noise[t][\alpha]\right) \right).
\end{align}
Finally, we can write down the full state and covariance update equations of the RI-EKF using the derived linear update equation and the theory of Kalman filtering~\citep{anderson1979optimal,bar2001estimation} as
\begin{align}
\XE[t][+] = \text{exp}\left( \K[t] \boldsymbol{\Pi} \left( \XE[t] \Y[t] \right)  \right) \XE[t] 
,\quad 
\P[t][+] = (\I - \K[t] \H[t]) \P[t],
\end{align}
where the gain $\K[t]$ is computed using
\begin{align*}
\S[t] = \H[t] \P[t] \H[t][\transpose] + \hat{\N}_t
\qquad
\K[t] = \P[t] \H[t][\transpose] \S[t][-1]
\end{align*}
and from  \eqref{eq:linear_update}, the matrices $\H[t]$ and $\hat{\N}_t$ are given by
\begin{equation*}
\begin{split}
\H[t] &= 
\begin{bmatrix}
\zeros[3,3] & \zeros[3,3] & -\I & \I 
\end{bmatrix}, \\
\hat{\N}_t &= \hat{\R}_t \; \J[v](\encodersM[t]) \; \text{Cov}(\noise[t][\alpha]) \; \J[v][\transpose](\encodersM[t]) \; \hat{\R}_t^\transpose.
\end{split}
\end{equation*}


\subsection{Observability Analysis}
Because the error dynamics are log-linear (c.f., Theorem~\ref{theorem:log_linear_error}), we can determine the unobservable states of the filter without having to perform a nonlinear observability analysis \cite{barrau2015non}. Noting that the linear error dynamics matrix in our case is time-invariant and nilpotent (with a degree of 3), the discrete-time state transition matrix is a polynomial in $\A[t]$,
\begin{equation*}
\boldsymbol{\Phi} = \text{exp}_m(\A[t] \Delta t) = 
\begin{bmatrix}
\I & \zeros & \zeros & \zeros \\
\vectorToSkew[\g] \Delta t & \I & \zeros & \zeros \\
\dfrac{1}{2} \vectorToSkew[\g] \Delta t^2 & \I \Delta t& \I & \zeros \\
\zeros & \zeros & \zeros & \I \\
\end{bmatrix}.
\end{equation*}
It follows that the discrete-time observability matrix is
\begin{equation*}
\mathcal{O} = 
\begin{bmatrix}
\H \\
\H \boldsymbol{\Phi} \\
\H \boldsymbol{\Phi}^2 \\
\vdots
\end{bmatrix}
=
\begin{bmatrix}
\zeros & \zeros & -\I & \I \\
- \dfrac{1}{2} \vectorToSkew[\g] \Delta t^2 & -\I \Delta t & -\I & \I \\
-2 \vectorToSkew[\g] \Delta t^2 & -2 \I \Delta t^2 & -\I & \I \\
\vdots & \vdots & \vdots & \vdots
\end{bmatrix}.
\end{equation*}
The last six columns (i.e., two matrix columns) of the observability matrix are clearly linearly dependent, which indicates the absolute position of the robot is unobservable. In addition, since the gravity vector only has a $z$ component, the third column of $\mathcal{O}$ is all zeros. Therefore, a rotation about the gravity vector (yaw) is also unobservable. This linear observability analysis agrees with the nonlinear observability results of \cite{bloesch2013state}, albeit with much less computation. Furthermore, as the error dynamics do not depend on the estimated state, there is no chance of the linearization spuriously increasing the numerical rank of the observability matrix \cite{barrau2015non}. This latter effect was previously known and studied in~\cite{bloesch2013state}, and in order to resolve this problem, an observability-constrained EKF~\cite{huang2010observability} was developed. In our proposed framework, by default, the discrete RI-EKF has the same unobservable states as the underlying nonlinear system; hence, the developed discrete RI-EKF inherently solves this problem.

\subsection{Comparison to Quaternion-Based EKF} \label{sec:sim}
\noindent

To investigate potential benefits or drawbacks of the proposed filter, we compare its response to initialization errors against that of a state-of-the-art quaternion-based EKF (Q-EKF), similar to those described in \cite{bloesch2013state,rotella2014state}. A dynamic \textit{simulation} of a Cassie-series bipedal robot (described in Section \ref{sec:experimental_results}) was performed in which the robot slowly walked forward after a small drop, accelerating from $0.0$ to $0.3~\m/\sec$. The simulated measurements were corrupted by additive white Gaussian noise, which are specified in Table \ref{tab:params} along with the initial state covariance values. The same values were used in both simulation and experimental evaluations of the proposed filter. The IMU bias estimation was turned off for these simulations. The simulation environment  models ground contact forces with a linear force law (having a stiffness and damping term) and a Coulomb friction model.


To compare the convergence properties of the two filters, 100 simulations of each filter were performed using the same measurements, noise statistics, initial covariance, and various random initial orientations and velocities. The initial Euler angle estimates were sampled uniformly from $-30\deg$ to $30\deg$. The initial velocity estimates were sampled uniformly from $-1.0~\m/\sec$ to $1.0~\m/\sec$. The pitch and roll estimates as well as the (body frame) velocity estimates for both filters are shown in Figure~\ref{fig:sim_comparison}. Although both filters converge for this set of initial conditions, the proposed RI-EKF converges considerably faster than the standard quaternion-based EKF. 

%% file: bias.tex
\section{Including IMU Biases}
\label{sec:bias}
Implementation of an IMU-based state estimator on hardware typically requires modeling additional states, such as gyroscope and accelerometer biases. Unfortunately, as noted in \cite{barrau2015non}, there is no Lie group that includes the bias terms while also having the dynamics satisfy the group affine property \eqref{eq:group_affine}. Even though many of the theoretical properties of the RI-EKF will no longer hold, it is possible to design an ``Imperfect InEKF'' that still outperforms the standard EKF \cite{barrau2015non}.

\vspace{-.2cm}
\subsection{State Representation}
The IMU biases are slowly varying signals that corrupt the measurements in an additive way:
\begin{alignat*}{2}
\angularVelocityM[t] &= \angularVelocity[t] + \b[t][g] + \noise[t][g], \qquad &&\noise[t][g] \sim \mathcal{N}(\zeros[3,1], \Cov[][g]) \\
\accelerationM[t] &= \acceleration[t] + \b[t][a] + \noise[t][a], \qquad &&\noise[t][a] \sim \mathcal{N}(\zeros[3,1], \Cov[][a]). 
\end{alignat*}
These biases form a parameter vector that needs to be accurately estimated as part of the RI-EKF state, 
\begin{equation}
\begin{split}
\params[t] &\triangleq 
\begin{bmatrix}
\gyroscopeBias(t) \\
\accelerometerBias(t) \\
\end{bmatrix} 
\triangleq 
\begin{bmatrix}
\gyroscopeBias[t] \\
\accelerometerBias[t] \\
\end{bmatrix} \in \realnumbers^6.
\end{split}
\end{equation}
The model's state now becomes a tuple of our original matrix Lie group and the parameter vector, $(\X[t], \params[t]) \in \mathcal{G} \times \realnumbers^6$. The augmented right-invariant error is now defined as
\begin{equation} \label{eq:augmented_error}
\e[t][r] \triangleq (\XE[t] \X[t][-1], \paramsE[t] - \params[t]) \triangleq (\groupError[t][r], \paramError[t]).
\end{equation}
Written explicitly, the right-invariant error is
\begin{equation*}
\resizebox{\hsize}{!}{$
\groupError[t][r] = 
\begin{bmatrix}
\RE[t] \R[t][\transpose] 
& \vE[t] - \RE[t] \R[t][\transpose] \v[t] 
& \pE[t] - \RE[t] \R[t][\transpose] \p[t] 
& \dE[t] - \RE[t] \R[t][\transpose] \d[t] \\
\zeros[1,3] & 1 & 0 & 0 \\
\zeros[1,3] & 0 & 1 & 0 \\
\zeros[1,3] & 0 & 0 & 1 \\
\end{bmatrix},$} 
\end{equation*}
while the parameter vector error is defined by
\begin{equation*}
\paramError[t] = 
\begin{bmatrix}
\gyroscopeBiasE[t] - \gyroscopeBias[t] \\
\accelerometerBiasE[t] - \accelerometerBias[t] \\
\end{bmatrix} 
\triangleq
\begin{bmatrix}
\paramError[t][g] \\
\paramError[t][a] 
\end{bmatrix}.
\end{equation*}
As detailed in \cite{barrau2015non}, the linearized process and measurement models will have a block structure: 
\begin{equation} \label{eq:block_matrix_structure}
\A[t] = 
\begin{bmatrix}
\A[\X] & \A[\X , \params] \\
\zeros & \A[\params]
\end{bmatrix}
,\quad
\H[t] = 
\begin{bmatrix}
\H[\X] & \H[\params]
\end{bmatrix}.
\end{equation}

\subsection{System Dynamics}
The deterministic system dynamics now depend on both the inputs, $\u[t]$, and the parameters, $\params[t]$:
\begin{equation*}
\resizebox{\hsize}{!}{$
f_{(\params[t],\u[t])}(\X[t]) = 
\begin{bmatrix}
  \R[t]\vectorToSkew[\angularVelocityM[t] - \gyroscopeBias[t]] 
& \R[t](\accelerationM[t] - \accelerometerBias[t]) + \g  
& \v[t]
& \zeros[3,1] \\
\zeros[1,3] & 0 & 0 & 0  \\
\zeros[1,3] & 0 & 0 & 0  \\
\zeros[1,3] & 0 & 0 & 0  \\
\end{bmatrix}.$}
\end{equation*}
The IMU bias dynamics are modeled using the typical ``Brownian motion'' model, i.e., the derivatives are white Gaussian noise, to capture the slowly time-varying nature of these parameters.
\begin{equation}
\dot{\textbf{b}}^g_t = \noise[t][bg] 
, \quad
\dot{\textbf{b}}^a_t = \noise[t][ba]. 
\end{equation}
To compute the linearized error dynamics, the augmented right-invariant error \eqref{eq:augmented_error} is first differentiated with respect to time, 
\begin{equation}
\begin{split}
\dfrac{\mathrm{d}}{\mathrm{d}t} \e[t][r] &= \left( 
\dfrac{\mathrm{d}}{\mathrm{d}t} \groupError[t][r],
\begin{bmatrix}
\noise[t][bg] \\
\noise[t][ba] \\
\end{bmatrix}
\right).
\end{split}
\end{equation}
After carrying out the chain rule and making the first order approximation, \mbox{\small$\groupError[t][r] = \text{exp}(\tangentError[t]) \approx \I[d] + \vectorToAlgebra[\tangentError[t]]$}, the individual terms of the invariant error dynamics become:
\begin{equation} \label{eq:augmented_error_dynamics}
\begin{split}
\dfrac{\mathrm{d}}{\mathrm{d}t} \left( \RE[t] \R[t][\transpose] \right) & \approx \vectorToSkew[\RE[t] \left( \noise[t][g]-\paramError[t][g] \right)] \\
\dfrac{\mathrm{d}}{\mathrm{d}t} \left( \vE[t] - \RE[t] \R[t][\transpose] \v[t] \right) &\approx  \vectorToSkew[\g] \tangentError[t][R] + \RE[t](\noise[t][a]-\paramError[t][a]) \\
&+ \vectorToSkew[\vE[t]]  \RE[t] \left( \noise[t][g]-\paramError[t][g] \right) \\
\dfrac{\mathrm{d}}{\mathrm{d}t} \left( \pE[t] - \RE[t] \R[t][\transpose] \p[t] \right) &\approx \tangentError[t][v] + \vectorToSkew[\pE[t]] \RE[t] \left( \noise[t][g]-\paramError[t][g] \right)\\
\dfrac{\mathrm{d}}{\mathrm{d}t} \left( \dE[t] - \RE[t] \R[t][\transpose] \d[t] \right) &\approx \vectorToSkew[\dE[t]] \RE[t] \left( \noise[t][g]-\paramError[t][g] \right) \\ 
&+ \RE[t] \FK[R](\encodersM[t]) \noise[t][v]. 
\end{split}
\end{equation}
The augmented invariant error dynamics only depends on the estimated trajectory though the noise and bias errors, $\paramError[t]$ (this is expected because when there are no bias errors, there is no dependence on the estimated trajectory). A linear system can now be constructed from \eqref{eq:augmented_error_dynamics} to yield,
\begin{equation*}
\dfrac{\mathrm{d}}{\mathrm{d}t} \left(
\begin{bmatrix}
\tangentError[t] \\
\paramError[t]
\end{bmatrix} \right)
= \A[t] \begin{bmatrix}
\tangentError[t] \\
\paramError[t]
\end{bmatrix} +
\begin{bmatrix}
\Adjoint[\XE[t]] & \zeros[12,6] \\
\zeros[6,12] & \I[6]
\end{bmatrix}
\noise[t],
\end{equation*}
where the noise vector is defined by
$$\noise[t] \triangleq \vector[\noise[t][g],\, \noise[t][a],\, \zeros[3,1], \FK[R](\encodersM[t]) \noise[t][v], \noise[t][bg], \noise[t][ba]].$$

\subsection{Forward Kinematic Measurements}
The forward kinematics position measurement \eqref{eq:fk_measurement} does not depend on the IMU biases. Therefore, the $\H[t]$ matrix can simply be appended with zeros to account for the augmented variables. The linear update equation becomes
\begin{equation*}
\begin{bmatrix}
\tangentError[t][+] \\
\paramError[t][+]
\end{bmatrix} =
\begin{bmatrix}
\tangentError[t] \\
\paramError[t]
\end{bmatrix}
- 
\begin{bmatrix}
\K[t][\tangentError] \\
\K[t][\paramError]
\end{bmatrix}
\left( \H[t]
\begin{bmatrix}
\tangentError[t] \\
\paramError[t]
\end{bmatrix}
- \hat{\R}_t \left(\J[v](\encodersM[t]) \noise[t][\alpha]\right) \right).
  \end{equation*}

\subsection{Final Continuous RI-EKF Equations}
The final ``Imperfect'' RI-EKF equations that include IMU biases can now be written down. The estimated state tuple is predicted using the following set of differential equations:
\begin{equation*}
\left( \dfrac{\mathrm{d}}{\mathrm{d}t} \XE[t] \,, \dfrac{\mathrm{d}}{\mathrm{d}t} \paramsE[t] \right) = \left( f_{(\paramsE[t],\u[t])}(\XE[t]) \,, \zeros[6,1] \right).
\end{equation*}
The covariance of the augmented right invariant error dynamics is computed by solving the Riccati equation
\begin{equation*}
\dfrac{\mathrm{d}}{\mathrm{d}t} \P[t] = \A[t] \P[t] + \P[t] \A[t][\transpose] + \hat{\Q}_t,
\end{equation*}
where the matrices $\A[t]$ and $\hat{\Q}_t$ are now defined using \eqref{eq:augmented_error_dynamics},
\begin{equation*}
\begin{split}
\A[t] &= 
\begin{bmatrix}
\zeros & \zeros & \zeros & \zeros & -\RE[t] & \zeros \\
\vectorToSkew[\g] & \zeros & \zeros & \zeros & -\vectorToSkew[\vE[t]]\RE[t] & -\RE[t] \\
\zeros & \I & \zeros & \zeros &  -\vectorToSkew[\pE[t]]\RE[t] & \zeros \\
\zeros & \zeros & \zeros & \zeros & -(\dE[t])_\times \RE[t] & \zeros \\
\zeros & \zeros & \zeros & \zeros & \zeros & \zeros \\
\zeros & \zeros & \zeros & \zeros & \zeros & \zeros \\
\end{bmatrix} \\
\hat{\Q}_t &= 
\begin{bmatrix}
\Adjoint[\XE[t]] & \zeros[12,6] \\
\zeros[6,12] & \I[6]
\end{bmatrix}
\text{Cov}(\noise[t])
\begin{bmatrix}
\Adjoint[\XE[t]] & \zeros[12,6] \\
\zeros[6,12] & \I[6]
\end{bmatrix}^\transpose. \\ 
\end{split}
\end{equation*}
The estimated state tuple is corrected though the update equations
\begin{equation*}
\small
\left(\XE[t][+], \params[t][+]\right) = \left( \text{exp}\left( \K[t][\tangentError] \boldsymbol{\Pi} \left( \XE[t] \Y[t] \right)  \right) \XE[t]\, , \;\; \paramsE[t] + \K[t][\paramError] \boldsymbol{\Pi} \left( \XE[t] \Y[t] \right) \right),
\end{equation*}
where the gains $\K[t][\tangentError]$ and $\K[t][\paramError]$ are computed from
\begin{align*}
\small
\S[t] = \H[t] \P[t] \H[t][\transpose] + \hat{\N}_t
\qquad
\K[t] = \P[t] \H[t][\transpose] \S[t][-1],
\end{align*}
with the following measurement, output, and noise matrices,
\begin{equation}
\small
\begin{split}
\Y[t][\transpose] &=
\begin{bmatrix}
\FK[p]^\transpose(\encodersM[t]) & 0 & 1 & -1
\end{bmatrix}, \\
\nonumber \H[t] &= 
\begin{bmatrix}
\zeros & \zeros & -\I & \I & \zeros & \zeros
\end{bmatrix}, \\
\hat{\N}_t &= \hat{\R}_t \; \J[v](\encodersM[t]) \; \text{Cov}(\noise[t][\alpha]) \; \J[v][\transpose](\encodersM[t]) \; \hat{\R}_t^\transpose.
\end{split}
\end{equation}
As indicated in \cite{barrau2015non}, the matrices $\A[t]$ and $\H[t]$ have the block structure shown in \eqref{eq:block_matrix_structure}.

\subsection{Discretization}
The continuous dynamics can be discretized by assuming a zero-order hold on the inputs and performing Euler integration from $t_k$ to $t_{k+1}$. The discrete dynamics for the individual state elements becomes:
\begin{equation*}
\small
\begin{split}
\RE[k+1] &= \RE[k] \; \text{exp}\left((\angularVelocityM[k] - \gyroscopeBiasE[k])\Delta t \right) \\
\vE[k+1] &= \vE[k] + \RE[k] (\accelerationM[k] - \accelerometerBiasE[k]) \Delta t + \g \Delta t \\
\pE[k+1] &= \pE[k] + \vE[k] \Delta t + \dfrac{1}{2} \RE[k] (\accelerationM[k] - \accelerometerBiasE[k]) \Delta t^2 + \dfrac{1}{2} \g \Delta t^2 \\
\dE[k+1] &= \dE[k], \qquad  
\gyroscopeBiasE[k+1] = \gyroscopeBiasE[k], \qquad
\accelerometerBiasE[k+1] = \accelerometerBiasE[k],
\end{split}
\end{equation*}
where $\Delta t = t_{k+1} - t_{k}$ and $\text{exp}(\cdot)$ is the exponential map for $\SO(3)$. A first-order approximation can be used to simplify integration of the Riccati equation, resulting in the following discrete-time covariance propagation equation,
\begin{equation*}
\begin{split}
\P[k+1] &= \boldsymbol{\Phi} \, \P[k]\boldsymbol{\Phi}^\transpose + \hat{\Q}_k,
\end{split}
\end{equation*}
where the discrete-time linear process model is given by \mbox{$\boldsymbol{\Phi} \triangleq \text{exp}_m(\A[t] \Delta t)$}, and the approximated discrete noise matrix is given by \mbox{$\hat{\Q}_k  \approx\boldsymbol{\Phi} \, \hat{\Q}_t \, \boldsymbol{\Phi}^\transpose \Delta t$}.

%% file: contact_switch.tex
\section{Switching Contact Points and State Augmentation}
\label{sec:switchcontact}
Sections \ref{sec:riekf} and \ref{sec:bias} derived the equations for the RI-EKF under the assumption that the contact point is unchanging with time. However, for legged robots, contacts are discrete events that are created and broken as a robot navigates through the environment. Therefore, it is important to be able to conveniently add and remove contact-point states to and from the observer.

\subsection{Removing Contact Points}
To remove a previous contact point from the state, we marginalize the corresponding state variable by simply removing the corresponding column and row from the matrix Lie group. The corresponding elements of the covariance matrix are also eliminated. This can be done through a simple linear transformation. For example, if the robot is going from one contact to zero contacts, then the newly reduced covariance would be computed by
\begin{equation*}
\small
\begin{split}
\begin{bmatrix}
\tangentError[t][R] \\
\tangentError[t][v] \\
\tangentError[t][p] \\
\end{bmatrix} &=
\begin{bmatrix}
\I & \zeros & \zeros & \zeros \\
\zeros & \I & \zeros & \zeros \\
\zeros & \zeros & \I & \zeros \\
\end{bmatrix}
\begin{bmatrix}
\tangentError[t][R] \\
\tangentError[t][v] \\
\tangentError[t][p] \\
\tangentError[t][d] \\
\end{bmatrix} \triangleq \F[r] \, \tangentError[t] \\
&\implies \P[t][new] = \F[r] \, \P[t] \, \F[r][\transpose].
\end{split}
\end{equation*}

\begin{figure*}[t!]
  \centering
    \includegraphics[width=1\textwidth]{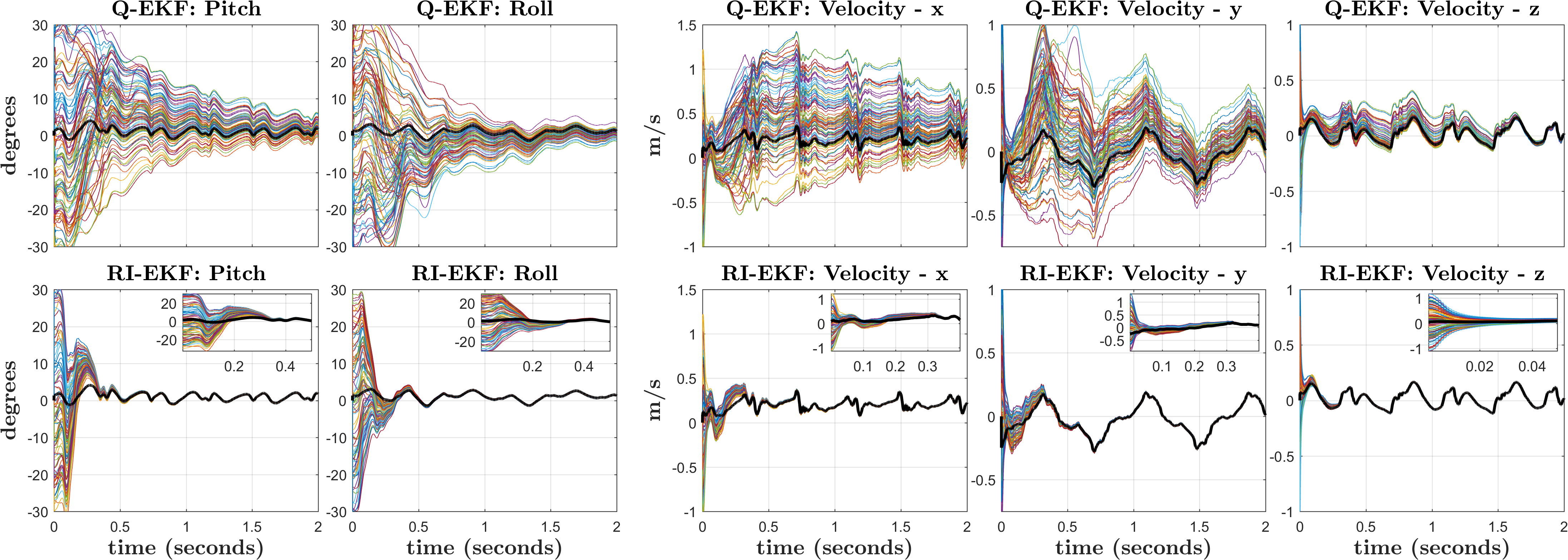}
	\caption{An experiment was performed where an actual Cassie-series robot slowly walked forwards at approximately $0.3~\m/\sec$. The noisy measurements came from the on-board IMU (VN-100) and the robot's joint encoders. The quaternion-based EKF (Q-EKF) and the proposed right-invariant EKF (RI-EKF) were run (off-line) 100 times using the same measurements, noise statistics, and initial covariance, but with random initial orientations and velocities. The black line represents the filter state estimates when initialized with a good estimate. The RI-EKF (bottom row) converges considerably faster than the Q-EKF (top row) for all observable states. Zoomed-in plots of the RI-EKF performance is provided in the top-right corner.}
\label{fig:exp_comparison_good_bias}
\squeezeup\squeezeup
\end{figure*}

\subsection{Adding Contact Points}
When the robot makes a new contact with the environment, the state and covariance matrices need to be augmented. Special attention needs to be given to initialize the mean and covariance for the new estimated contact point. For example, if the robot is going from zero contacts to one contact, the initial estimate is obtained though the forward kinematics relation
\begin{equation}
\small
\dE[t] = \pE[t] + \RE[t] \FK[p](\encodersM[t]). 
\end{equation}
In order to compute the new covariance, we need to look at the right-invariant error, 
\begin{equation*}
\small
\begin{split}
\dE[t] - \RE[t]\R[t][\transpose]\d[t] &= \pE[t] - \RE[t]\R[t][\transpose]\d[t] + \RE[t] \FK[p](\encodersM[t]) \\
\groupError[t][d] &= \pE[t] - \RE[t]\R[t][\transpose] \left( \p[t] + \R[t] \FK[p](\encodersM[t] - \noise[t][\alpha])  \right) + \RE[t] \FK[p](\encodersM[t]) \\
\groupError[t][d] &\approx \groupError[t][p] + \RE[t] \J[v](\encodersM[t]) \noise[t][\alpha] \\
\implies \tangentError[t][d] &= \tangentError[t][p] + \RE[t] \J[v](\encodersM[t]) \noise[t][\alpha].
\end{split}
\end{equation*}
Therefore, covariance augmentation can be done using the following linear map,
\begin{equation*}
\small
\begin{split}
\begin{bmatrix}
\tangentError[t][R] \\
\tangentError[t][v] \\
\tangentError[t][p] \\
\tangentError[t][d] \\
\end{bmatrix} &=
\begin{bmatrix}
\I & \zeros & \zeros \\
\zeros & \I & \zeros \\
\zeros & \zeros & \I \\
\zeros & \zeros & \I \\
\end{bmatrix}
\begin{bmatrix}
\tangentError[t][R] \\
\tangentError[t][v] \\
\tangentError[t][p] \\
\end{bmatrix} + 
\begin{bmatrix}
\zeros \\
\zeros \\
\zeros \\
\RE[t] \J[v](\encodersM[t]) \\
\end{bmatrix} \noise[t][\alpha] \\
\tangentError[t][\mathrm{new}] &\triangleq \F[a] \, \tangentError[t] + \G[t] \noise[t][\alpha] \\
\implies \P[t][\mathrm{new}] &= \F[a] \, \P[t] \, \F[a][\transpose] + \G[t] \, \text{Cov}(\noise[t][\alpha]) \, \G[t][\transpose]. 
\end{split}
\end{equation*}

%% file: experimental_results.tex
\section{Experimental Results on Cassie Robot}
\label{sec:experimental_results}

We now present an experimental evaluation of the proposed contact-aided RI-EKF observer using a 3D biped robot. The Cassie-series biped robot, shown in Figure~\ref{fig:cassie}, has 20 degrees of freedom coming from the body pose, 10 actuators, and 4 springs. The robot is equipped with an IMU along with 14 joint encoders that can measure all actuator and spring angles. The proposed and baseline algorithms (along with the robot's feedback controller) are implemented in MATLAB (Simulink Real-Time). The IMU (model VN-100) is located in the robot's torso and provides angular velocity and linear acceleration measurements at $800 \Hz$. The encoders provide joint angle measurements at $2000 \Hz$. The robot has two springs on each leg that are compressed when the robot is standing on the ground. The spring deflections are measured by encoders and are thresholded to serve as a binary contact sensor.

\subsection{Contact-aided Legged Odometry Experiment}
An experiment was performed where the robot walked forwards at approximately $0.3~\m/\sec$. The quaternion-based EKF (Q-EKF) and the proposed right-invariant EKF (RI-EKF) were run (off-line) 100 times using the same logged measurements, noise statistics, and initial covariance with random initial orientations and velocities. The noise statistics and initial covariance estimates are provided in Table~\ref{tab:params}. As with the simulation comparison presented in Section~\ref{sec:sim}, the initial mean estimate for the Euler angles were uniformly sampled from $-30\deg$ to $30\deg$ and the initial mean estimate for velocities were sampled uniformly from $-1.0~\m/\sec$ to $1.0~\m/\sec$. Bias estimation was turned on and the initial bias estimate was obtained from processing the IMU data when the robot was static. The pitch and roll estimates as well as the (body frame) velocity estimates for both filters are shown in Figure~\ref{fig:exp_comparison_good_bias}. The experimental results for comparing filter convergence matches those of the simulation. The proposed RI-EKF converges faster and more reliably in all 100 runs than the quaternion-based EKF; therefore, due to the convenience of initialization and reliability for tracking the developed \mbox{RI-EKF} is the preferred observer.

\begin{table}[t]
	\centering
    \caption{Experiment Noise Statistics and Initial Covariance}
    \resizebox{1\columnwidth}{!}{%
	{\renewcommand{\arraystretch}{1.0}%
    \begin{tabular}{cc}
      \begin{tabular}{l|l}
            \toprule
			Measurement	Type        & noise st. dev.\\
            \midrule
			Linear Acceleration	    & $0.04 ~\m/\sec^2$\\
			Angular Velocity        & $0.002 ~\rad/\sec$\\
			Accelerometer Bias      & $0.001 ~\m/\sec^2$\\
			Gyroscope Bias          & $0.001 ~\rad/\sec$\\
			Contact Linear Velocity & $0.05 ~\m/\sec$  \\
			Joint Encoders  & $1.0 ~\deg$  \\
            & \\
            \bottomrule
	  \end{tabular} &
      \begin{tabular}{l|l}
              \toprule
              State Element           & initial st. dev.\\
              \midrule
              Orientation	of IMU     & $30.0 ~\deg$ \\
              Velocity of IMU   	   & $1.0 ~\m/\sec$\\
              Position of IMU        & $0.1 ~\m$\\
              Position of Right Foot & $0.1 ~\m$\\
              Position of Left Foot  & $0.1 ~\m$  \\
              Gyroscope Bias         & $0.005 ~\rad/\sec$\\
              Accelerometer Bias     & $0.05 ~\m/\sec^2$\\
              \bottomrule
      \end{tabular}
    \end{tabular}
    }}
\label{tab:params}
\squeezeup\squeezeup
\end{table}

\subsection{Discussion}
When the state estimate is initialized close to the true value, the RI-EKF and Q-EKF have similar performance, because the linearization of the error dynamics accurately reflects the underlying nonlinear dynamics. However, when the state estimate is far from the true value, the simulation and experimental results show that RI-EKF consistently converges faster than the Q-EKF. The relatively poor performance of the Q-EKF is due to the error dynamics being linearized around the wrong operating point; therefore, the linear system does not accurately reflect the nonlinear dynamics. In addition, when bias estimation is turned off, the invariant error dynamics of the RI-EKF do not depend on the current state estimate. As a result, the linear error dynamics can be accurately used even when the current state estimate is far from its true value, leading to better performance over the Q-EKF. Although this theoretical advantage is lost when bias estimation is turned on, the experimental results (shown in Figure \ref{fig:exp_comparison_good_bias}) indicate that the RI-EKF still is the preferred observer due to less sensitivity to initialization. 